\documentclass[preprint, review,3p,times, 10pt, authoryear]{elsarticle}

\usepackage{amsmath,amsfonts}
\usepackage{algorithmic}
\usepackage{algorithm}
\usepackage{array}
\usepackage{booktabs}
\usepackage{pifont}
\newcommand{\xmark}{\ding{55}}  
\newcommand{\cmark}{\ding{51}}  

\usepackage{textcomp}
\usepackage{stfloats}
\usepackage{verbatim}
\usepackage{amssymb}
\usepackage{tablefootnote}
\usepackage{graphicx}
\usepackage{makecell}
\usepackage{multirow}
\usepackage{makecell}
\usepackage{amssymb}
\usepackage[hyphens]{url}
\usepackage{hyperref}
\hypersetup{colorlinks=true,breaklinks=true}
\usepackage{tikz}
\usepackage{pgfplots}
\usepackage{pgfplotstable}
\usepackage{subcaption}
\pgfplotsset{compat=1.17}
\usepgfplotslibrary{external}
\journal{Elsevier}

\begin{document}

\begin{frontmatter}

\title{CCoMAML: Efficient Cattle Identification Using Cooperative Model-Agnostic Meta-Learning}

\author[1,2,3]{Rabin Dulal}\ead{rdulal@csu.edu.au}
\author[1,2,3]{Lihong Zheng}\ead{lzheng@csu.edu.au}
\author[1,2,3]{Ashad Kabir}\ead{akabir@csu.edu.au}
\cortext[correspondingauthor]{Corresponding author: Rabin Dulal}
%
\affiliation[1]{organization={School of Computing, Mathematics and Engineering, Charles Sturt University}, city={Wagga Wagga}, state={NSW}, postcode={2678}, country={Australia}}
\affiliation[2]{organization={Gulbali Institute for Agriculture, Water and Environment, Charles Sturt University}, city={Wagga Wagga}, state={NSW}, postcode={2678},  country={Australia}}

\affiliation[3]{organization={Food Agility CRC Ltd}, city={Sydney}, state={NSW}, postcode={2000},  country={Australia}}
\begin{abstract}
Cattle identification is critical for efficient livestock farming management, currently reliant on radio-frequency identification (RFID) ear tags. However, RFID-based systems are prone to failure due to loss, damage, tampering, and vulnerability to external attacks. As a robust alternative, biometric identification using cattle muzzle patterns similar to human fingerprints has emerged as a promising solution. Deep learning techniques have demonstrated success in leveraging these unique patterns for accurate identification. But deep learning models face significant challenges, including limited data availability, disruptions during data collection, and dynamic herd compositions that require frequent model retraining. To address these limitations, this paper proposes a novel few-shot learning framework for real-time cattle identification using Cooperative Model-Agnostic Meta-Learning (CCoMAML) with Multi-Head Attention Feature Fusion (MHAFF) as a feature extractor model. This model offers great model adaptability to new data through efficient learning from few data samples without retraining. The proposed approach has been rigorously evaluated against current state-of-the-art few-shot learning techniques applied in cattle identification. Comprehensive experimental results demonstrate that our proposed CCoMAML with MHAFF has superior cattle identification performance with 98.46\% and 97.91\% F1 scores. 
\end{abstract}

\end{frontmatter}

\section{Introduction}
The growing emphasis on biosecurity and food safety has led to a heightened need for reliable cattle traceability, which relies heavily on accurate identification methods. Cattle identification techniques are generally divided into three categories: traditional, electronic, and modern vision-based approaches~\citep{LR1awad2016classical, hossain2022systematic}. Traditional methods include ear tattoos, branding, ear tags, and ear notching~\citep{CIOruiz2011role, LR1awad2016classical, INneary2002methods}. Ear tags, for instance, usually contain a unique identification number along with details such as breed, sex, and birth date. These tags are attached to the animal’s ear using a specific applicator. Electronic identification is more efficient than traditional methods and is widely adopted today. In Australia, for example, cattle are tracked using the National Livestock Identification System (NLIS), which employs radio-frequency identification (RFID) tags~\citep{NLIS}. While RFID tags enhance the traceability process, they are prone to issues such as tampering, loss, damage, and alteration~\citep{CIOruiz2011role, INneary2002methods, LR1awad2016classical, andrew2019visual, ruiz2011role, wang2010rfid}. Biometric methods have emerged as a promising alternative to RFID tags. These include techniques based on iris recognition, retina scanning, DNA analysis, coat patterns, and muzzle patterns. However, capturing images of the retina and iris is labor-intensive, and DNA-based methods demand specialized equipment, laboratory facilities, and time. Similarly, coat pattern recognition is effective only for cattle with unique markings and is unsuitable for those with uniform coloration.

Muzzle patterns serve as a distinctive form of biometric identification for cattle, similar to how fingerprints function in humans~\citep{kumar2020cattle}. These patterns consist of unique arrangements of beads and ridges on the muzzle skin~\citep{baranov1993breed}. Although the size of these patterns may increase as the animal matures, the specific structure remains consistent over time~\citep{petersen1922identification, LR1awad2016classical}. One of the key advantages of using muzzle patterns is that they can be easily captured through photography, eliminating the need for complex procedures, specialized tools, or laboratory settings~\citep{AID32kumar2018deep, awad2019bag, kumar2017muzzle}. This ease of use, combined with their uniqueness, makes muzzle patterns a practical and reliable option for biometric cattle identification~\citep{CIOruiz2011role, INneary2002methods, LR1awad2016classical, andrew2019visual, ruiz2011role, wang2010rfid}.

In recent years, deep learning has revolutionized computer vision by enabling automatic extraction of meaningful features directly from raw image data~\citep{lecun2015deep, hou2015convolutional, sharif2014cnn}. Two of the most prominent deep learning models used in visual tasks, including cattle identification, are Convolutional Neural Networks (CNNs) and transformer-based architectures~\citep{vaswani2017attention}. CNNs have shown notable success in cattle identification using muzzle images, delivering strong and consistent results~\citep{AID32kumar2018deep, vincent2008extracting, bengio2009learning, bengio2013representation, shojaeipour2021automated, li2022individual, lee2023identification}. More recently, transformer models such as the Vision Transformer (ViT)\citep{dosovitskiy2020image} and the Swin Transformer\citep{liu2021swin} have also gained attention in this field, demonstrating encouraging outcomes and offering new research directions~\citep{bergman2024biometric, wu2021body, guo2023vision, zhang2023high}. CNNs are particularly efficient in capturing local features, but their limited receptive fields make it difficult to model long-range relationships~\citep{yang2024feature, naseer2021intriguing}. On the other hand, transformers can model global dependencies through self-attention mechanisms but lack inherent inductive biases such as locality and shift invariance, which are important for detailed local feature representation~\citep{dosovitskiy2020image, han2022survey, khan2022transformers}. As a result, relying exclusively on either architecture may hinder performance and generalization~\citep{mogan2024ensemble}. To address these challenges, recent studies have proposed hybrid models that combine CNNs and transformers, leveraging their complementary strengths to enhance feature learning and increase model robustness~\citep{dulal2025mhaff, yang2024feature, li2023x, ojala1994performance, li2023combining, du2022individual, wan2023sheep, fu2022lightweight, hu2020cow, weng2022cattle}.

Deep learning in cattle identification faces key challenges, such as limited data availability~\citep{hossain2022systematic} and the frequent need to retrain models due to changing herd compositions~\citep{xu2024few, zhang2023siamese, bakhshayeshi2023intelligence, porto2024new, xu2022cow}. Deep learning models typically require large, diverse datasets for effective training; however, collecting extensive cattle data is time-consuming, requires more resources, and can be disruptive to the animals and their natural environment. Moreover, cattle herds are dynamic—cattle are frequently added or removed, making it difficult to maintain consistent and up-to-date identification models. To address limited data availability, several techniques have been employed, including data augmentation and transfer learning. Data augmentation involves generating synthetic data from existing samples through manual image transformations such as translation, flipping, shearing, scaling, reflection, cropping, and rotation. More advanced augmentation techniques like mosaic augmentation~\citep{bochkovskiy2020yolov4}, mixup~\citep{zhang2017mixup}, and cutout~\citep{devries2017improved} are also used to create diverse combinations of training samples and increase dataset size. However, data augmentation has its limitations: augmentation policies are often tailor-made for specific datasets, and these handcrafted rules generally do not generalize well to datasets from different domains~\citep{wang2020generalizing}. Transfer learning is another powerful approach, especially beneficial when working with small datasets. Instead of training a model from scratch, which typically requires large labeled datasets to prevent overfitting, transfer learning uses pretrained weights from models trained on large-scale datasets and fine-tunes them for new classes with limited data~\citep{torrey2010transfer}. Nonetheless, transfer learning is not always effective~\citep{dumoulin2021comparing}. In the fine-tuning setting, the entire model is updated using training data from the downstream task, which may lead to reduced performance~\citep{abnar2021exploring}. Furthermore, the dynamic nature of cattle herds presents another limitation. Deep learning models are typically trained on a fixed number of classes, making them unsuitable for scenarios where the herd size changes. When a new animal is added to the herd, the model must be retrained from scratch with updated data that includes the new individual. This process is tedious, time-consuming, and inefficient~\citep{xu2024few}. 

Few-shot learning (FSL) is an effective approach for addressing the challenge of training models in data-scarce environments that is not solved by data augmentation and transfer learning~\citep{dumoulin2021comparing, abnar2021exploring, wang2020generalizing, vettoruzzo2024advances}. It emulates the human ability to learn and generalize from only a few examples. Unlike traditional supervised learning, which requires large amounts of labeled data and extensive training, FSL is specifically designed to perform well with minimal data. To address these issues, this study proposes a novel few-shot learning framework for real-time cattle identification using a meta-learning approach. Specifically, we introduce CCoMAML—a Cooperative Model-Agnostic Meta-Learning~\citep{shin2024cooperative} method that incorporates Multi-Head Attention Feature Fusion (MHAFF)~\citep{dulal2025mhaff} as its backbone. This framework enables rapid adaptation to new cattle with minimal data, eliminating the need for full retraining. This work provides a comprehensive design and implementation framework for meta-learning in livestock management, contributing to the development of a robust and flexible cattle biometric system. The major contributions of this paper are as below:

\begin{itemize}
    \item A new approach called CCoMAML was proposed for individual cow identification, specifically designed to work effectively when only a limited number of muzzle images are available. The proposed method also handles dynamic herd sizes of cattle without training from scratch. 
    \item A custom CNN-based co-learner was proposed to provide better generalization ability by introducing learnable noise in gradient modification.
    \item A comparative evaluation was conducted involving 14 state-of-the-art few-shot learning (FSL) methods alongside the proposed CCoMAML model. To the best of our knowledge, this is the first study to systematically compare FSL techniques for cattle identification. Extensive experiments were carried out to validate the effectiveness of the proposed approach. Moreover, the proposed CCoMAML is also validated on two publicly available benchmark datasets. 
\end{itemize}

The remainder of this article is organized as follows. Section~\ref{background} introduces basic knowledge of the FSL methods used in this research. Section~\ref{related_works} reviews the related works, including deep learning and few-shot learning-based approaches for cattle identification. Section~\ref{problem_formulation} presents problem formulation. Section~\ref{methodology} outlines the proposed methodology, detailing the datasets used, the selection of the base model, and the chosen few-shot learning method. Section~\ref{experiments and results} presents the experimental setup and results. Finally, Section~\ref{conclusion} concludes the paper with a summary of findings and suggestions for future work.

\section{Background}
\label{background}
Few-shot learning (FSL) is a technique that enables models to learn and generalize from only a small number of labeled examples, mimicking the human ability to recognize new concepts with minimal data. FSL can be divided into two main types: metric-based and optimization-based methods. This section provides a basic understanding of metric and optimization-based FSL algorithms. 

\subsection{Metric Based FSL}
Metric-based methods learn how to compare new data with a few known examples using distance measures like Euclidean or cosine similarity. They work by placing similar items close together in a learned space and predicting based on which known example is most similar. This section provides background information on metric-based FSL algorithms, including Siamese Neural Network~\citep{bromley1993signature}, Prototypical Networks (Protonet)~\citep{snell2017prototypical}, Matching Networks~\citep{vinyals2016matching}, and Relation Networks~\citep{sung2018learning}. 

\subsection{Optimization Based FSL}
Optimization-based methods focus on making the model quickly learn a new task using only a few training examples. These methods train the model to adapt its parameters in just a few steps to perform well even with minimal data. This section provides background information of optimization-based FSL algorithms, like Model-Agnostic Meta-Learning (MAML)~\citep{finn2017maml}, Reptile~\citep{nichol2018reptile}, Simple Neural AttentIve Learner (SNAIL)~\citep{mishra2017simple}, Meta-SGD~\citep{li2017meta}, First Order MAML (FOMAML)~\citep{finn2017maml}, MAML++~\citep{antoniou2018train}, Almost-No-Inner-Loop algorithm (ANIL)~\citep{raghu2019rapid}, Body Only update in Inner Loop (BOIL)~\citep{oh2020boil}, Latent Embedding Optimization (LEO)~\citep{rusu2018meta}, and Cooperative Meta Learning (CML)~\citep{shin2024cooperative}. 

\section{Related Works}
\label{related_works}
This section is divided into two parts: the first part presents recent studies on cattle identification using deep learning (DL) approaches, while the second part focuses on approaches based on few-shot learning (FSL).

\subsection{Cattle Identification using DL}
Deep learning models such as ResNet, VGG, InceptionNet, Vision Transformer (ViT), MobileNet, EfficientNet, and Swin Transformer have been instrumental in cattle identification using muzzle images. This section presents existing studies that applied DL models for cattle identification using muzzle images. These studies applied transfer learning by leveraging deep learning models pre-trained on ImageNet.

\citet{shojaeipour2021automated} proposed a cattle identification method using ResNet, demonstrating high performance. Similarly, \citet{bezen2020computer, salau2020instance, liu2020video, xu2020automated, qiao2019cattle, xiao2022cow} employed ResNet for its strong feature extraction capabilities. Building on this, \citet{kimani2023cattle} used Wide ResNet to further enhance identification performance. VGG has also been applied effectively in cattle identification. \citet{li2022individual} reported strong performance using VGG among various deep learning models. Several other studies~\citep{li2022generalized, guzhva2018now, wu2021using, li2019deep, wang2020cattle, rivas2018detection} have also adopted VGG-based approaches. \citet{kimani2023cattle} compared VGG and Wide ResNet, with Wide ResNet showing better results. InceptionNet has shown strong performance in cattle identification, with studies leveraging its multi-scale feature extraction capabilities~\citep{qiao2019individual, qiao2021automated, ren2021tracking, porter2021feasibility}. EfficientNet has also been used effectively, particularly for Hanwoo cattle, due to its balance of accuracy and efficiency through compound scaling~\citep{lee2023identification, saar2022machine, yin2020using}. Lightweight models like MobileNet have enabled portable identification systems suitable for field deployment~\citep{li2021individual, hou2021cow, zin2020automatic, xudong2020automatic}. ViT has emerged as a promising alternative to CNNs by capturing global context in muzzle images~\citep{bergman2024biometric, wu2021body, guo2023vision, zhang2023high}. Further extending transformer-based approaches, the Swin Transformer has been explored for its efficient handling of high-resolution images~\citep{lu2022recognition, zhao2023research, zhong2023method}.

Feature fusion has shown clear advantages over using individual DL models in cattle identification. By combining features from different models or layers, it captures richer representations that improve accuracy. Common fusion methods include addition and concatenation, as seen in studies combining CNN and transformer features~\citep{li2023combining}, or fusing local and global features~\citep{du2022individual}. Recently, Multi-Head Attention-based Feature Fusion (MHAFF) has demonstrated superior performance by effectively integrating CNN and transformer features, outperforming both traditional fusion techniques and standalone models~\citep{dulal2025mhaff}.

However, deep learning models typically require large amounts of data, which poses challenges in collecting numerous muzzle images of cattle. The process is time-consuming, and frequent image capturing can disturb cattle in their natural habitat.

\subsection{Cattle Identification using FSL}
Few-shot learning methods are well-suited for cattle identification tasks requiring generalization to new and unseen data, even with a limited number of samples, such as five muzzle images per individual. This subsection provides recent and relevant research based on a few shots of learning to identify cattle. 

In few-shot learning, distance metric learning, like a Siamese network~\citep{bromley1993signature}, is pivotal for enabling models to generalize from limited labeled examples effectively. The core idea is to learn an embedding or feature space where data points from the same class are close together while those from different classes are farther apart. This structure allows the model to classify new, unseen cattle by measuring distances in this learned space. ~\citet{andrew2021visual} proposed a method to identify individual cattle, both seen and unseen, without requiring extensive manual labeling or re-training of a closed-set classifier. Their approach leverages metric learning to map input images into a class-distinctive latent space. In this space, embeddings of images belonging to the same individual naturally cluster together, effectively reducing the input dimensionality from a large image matrix to a compact vector. The distances in this latent space directly represent input similarity, enabling lightweight clustering algorithms, such as k-Nearest Neighbors, for classification. This method facilitates efficient and scalable cattle identification by focusing on clustering feature embeddings rather than traditional classification models. The author combined reciprocal triplet loss~\citep{masullo2019goes} with softmax to cluster similar and different cattle. Similar to this idea, ~\citet{chen2021lightweight} proposed a cow identification model using a triplet loss~\citep{hermans2017defense} as a distance metric to learn the differences between the different cattle features. \citet{zhang2022unsupervised} implemented distance metric learning using raw and deep distance metrics. The raw distance is determined by comparing pixel-level differences between frames and support images, identifying the closest match based on minimal visual differences. While effective for essential similarity detection, this approach may miss deeper semantic relationships. To address this, the authors introduced a deep distance metric using features extracted from a pre-trained model. These high-level features provide a richer representation of features, capturing intricate details beyond raw pixel values. The deep distance evaluates the similarity in the extracted feature spaces. Finally, the method combines raw and deep distances, balancing their contributions to enhance accuracy. ~\citet{qiao2022one} proposes a one-shot learning-based segmentation method with pseudo-labeling to reduce the need for extensive pixel-labeled data when segmenting animals in videos. Using only one labeled frame, the approach achieved superior performance on a challenging cattle video dataset compared to state-of-the-art methods. This technique offers a promising solution for efficient animal monitoring in smart farming applications.

~\citet{xu2022cow} proposed an enhanced Siamese Neural Network. The approach combines Dense Blocks (DB) and Capsule Networks to enhance feature extraction and retain spatial information. Image pairs are processed through a Siamese Neural Network to generate bivariate features, which are analyzed for correlation to achieve recognition. The model generalizes well to unfamiliar cows. ~\citet{porto2024new} introduced a two-phase training strategy for a Siamese neural network. In the first phase, contrastive loss is used to train the network to optimize the distance between the feature representations of two input images, focusing on similarity recognition. In the second phase, the network incorporates a classification head and is optimized for classification tasks. ~\citet{bakhshayeshi2023intelligence} compared two loss functions using a Siamese network as a few-shot learning algorithm for cattle reidentification on farms. They found that the triplet loss generalized better to test data, increasing cattle identification accuracy. ~\citet{zhang2023siamese} used an improved Siamese network for the identification of cattle by using head images. 

~\citet{meng2024few} proposed a few-shot learning method for cattle identification using the Prototypical Network, or ProtoNet. The approach employs a shared branch for extracting common features and a private branch for capturing unique features. This design addresses the challenge of small inter-class variations in the dataset. Comparative experiments demonstrate that the method achieves high accuracy in cow identification using limited sample images, outperforming existing algorithms.  

~\citet{xu2024few} proposed a method based on the Model-Agnostic Meta-Learning (MAML) framework to improve the accuracy of cattle identification under few-shot conditions, enabling quick adaptation to changes in individual cows. Additionally, an autoencoder was integrated to help the model learn more generalized features by combining supervised and unsupervised learning techniques. Experimental results demonstrated the effectiveness of this approach and identification using only five samples per cow. Moreover, ~\citet{garcia2024meta} proposed a cattle identification and weight prediction method based on the MAML framework, which provides an impressive application. Interestingly, ~\citet{moon2024task, poutaraud2023unsupervised} proposed a method to identify the cattle using the MAML framework based on the sound of the individual cattle.

Table~\ref{fsl_compare_other} summarizes aforementioned studies based on FSL for cattle identification. The summary table presents recent studies, their region of interest (RoI), and the best-performing few-shot learning (FSL) approach. To the best of our knowledge, existing FSL-based cattle identification studies have primarily focused on visual features such as coat patterns, facial structures, and teat characteristics. However, none have explored the use of muzzle images. This research is the first to introduce muzzle-based cattle identification within a few-shot learning framework, filling the gap with a comprehensive analysis of 14 state-of-the-art FSL methods, followed by proposing a new method called CCoMAML. 
\begin{table}[!t]
    \centering
    \caption{Summary of FSL-based studies in cattle identification.}
    \begin{tabular}{lcll}
    \hline
        Study & RoI & Best Method & Strategy\\ \hline\hline
        \citet{andrew2021visual} & Coat & Deep Metric Learning &Metric\\
        \citet{chen2021lightweight} & Face & Deep Metric Learning & Metric\\
        \citet{zhang2022unsupervised} & Teat & Distance Metric Learning &Metric\\
        \citet{qiao2022one} & Coat & Xception-FCN &Metric\\
        \citep{xu2022cow} & Face & Siamese &Metric\\
        \citep{meng2024few} & Face & Protonet &Metric\\
        \citep{xu2024few} & Body & MAML &Optimization\\
        \textbf{This work} & \textbf{Muzzle} & \textbf{CCoMAML} &Optimization\\
        \hline
    \end{tabular}
    \label{fsl_compare_other}
\end{table}

\section{Problem Formulation}
\label{problem_formulation}
Let us start with a group \(A\) of \(N_A\) number of cattle; several muzzle images are collected for each cattle. A deep learning classification model \(f_\theta^A\) can be used to represent cattle identity. This model learns the ability to extract embedding or latent features from raw muzzle images and can predict an unseen muzzle image. \(\theta\) represents the knowledge learned by the model \(f_\theta^A\) for the cattle group A.

The challenge in deep learning-based cattle identification systems is that the prior cattle classification model \(f_\theta^A\) cannot be directly adapted to a new task of identifying other unseen cattle. The new identification task arises in scenarios when an additional number of $i$ cattle are introduced to the group \(A\). Therefore, a new dataset must be built for added classes, and the previously trained model \(f_\theta^A\) needs to be retrained for an \(N_A\)+ $i$  cattle classification model \(f_\theta^{A'}\). 

In real-world applications, this kind of scenario often happens. Such a time-consuming and impractical process is not feasible to address some challenging situations, such as animal trade or health emergencies, where quick decision-making for identification is essential. Hence, an adaptable solution is critical that allows the model to be quickly updated without requiring a complete retraining process.

Moreover, for a new group \(B\) with \(N_B\) number of cattle in different breeds belonging to different farms, the pre-training identification model \(f_\theta^A\) developed for group \(A\) cannot be directly applied to group \(B\). This is because a deep learning-based cattle identification model relies heavily on specific training data characteristics.

As a result, when the pre-trained cattle identification model is applied to a new, unknown dataset, the model needs to be retrained entirely from scratch. This static nature of conventional deep learning methods makes them impractical for dynamic and varied scenarios, such as identifying cattle across farms with different breeds, environmental conditions, and data collection processes. 

In those cases, it needs to design a new way with a self-learning ability to handle the general domain of \(N\)-cattle identification problems (N is the classification number of cattle), especially with limited examples (shots) of cattle images. This problem, in contrast to the existing deep learning approaches, which require large labeled training samples, can be solved through meta-learning, commonly known as the \(N\)-cattle, few-shot learning problem~\citep{hospedales2021meta, vettoruzzo2024advances, gharoun2024meta}.

Meta-learning, or "learning to learn," is a technique where models are trained to quickly adapt to new tasks with minimal data by leveraging experience from a variety of previous tasks. It focuses on learning generalizable knowledge or strategies (such as initialization, optimization, or similarity metrics) that improve learning efficiency in new, unseen scenarios. Meta-learning is especially useful for few-shot learning problems, where the goal is to enable models to accurately learn from only a few labeled examples by rapidly adapting using the knowledge gained from related tasks. It generalizes well for predicting unseen cattle images~\citep{wang2023few} and the dynamic size of the cattle herd~\citep{zhou2021fewshot}, making it a suitable choice for addressing the specific challenges of cattle identification, reducing the need for extensive retraining~\citep{zhang2023survey, hospedales2021meta, vettoruzzo2024advances, gharoun2024meta}. Let \(\theta\) denote the meta-knowledge learned during meta-training, and \(\phi\) denote the task-specific knowledge parameters obtained through training on task-associated labeled data. A task refers to a specific learning problem or mini-dataset sampled from a larger distribution. Each task comprises a small amount of training data (support set) and evaluation data (query set). Specifically, for the application of cattle identification task, \(\mathcal{T}\), the goal is to learn a meta-function, \(f_\theta(\mathcal{D}_\mathcal{T})\) for \(p(\mathcal{T})\) 
where \(\mathcal{D}_\mathcal{T}\) represents the dataset associated with the task consisting of muzzle images and their corresponding labels.\(\mathcal{T}\) and \(p(\mathcal{T})\) represents the probability distribution over the task 
\(\mathcal{T}\). 

Even though \(f_\theta\) is trained on one dataset, it is not specific to that dataset. It learns a general, reusable initialization parameter that performs well on new, unseen data. For the determined number of cattle identification tasks \(\mathcal{T}_j\), where $j$ represents a specific task, the meta-learning process can be approximated as:
\begin{equation}
\label{problem_formulation_equation}
f_\theta(\mathcal{D}_\mathcal{T}, \mathcal{L}_\mathcal{D_\mathcal{T}}) \rightarrow f_\phi(\mathcal{D}_{\mathcal{T}_j}, \mathcal{L}_{\mathcal{D}_{\mathcal{T}_j}}).
\end{equation}

The loss function used to train the model on a specific dataset is represented as \(\mathcal{L}_\mathcal{D_\mathcal{T}}\). 

The task-specific model, parameterized by \(\phi\), is represented as \(f_\phi\), optimized for a particular task-focused dataset. The meta-function \(f_\theta\), parameterized by \(\theta\), learns meta-knowledge during meta-training and produces better-initialized task-specific models.

\section{Methodology}
\label{methodology}
\begin{figure}[ht]
    \centering
    \includegraphics[width=0.60\linewidth]{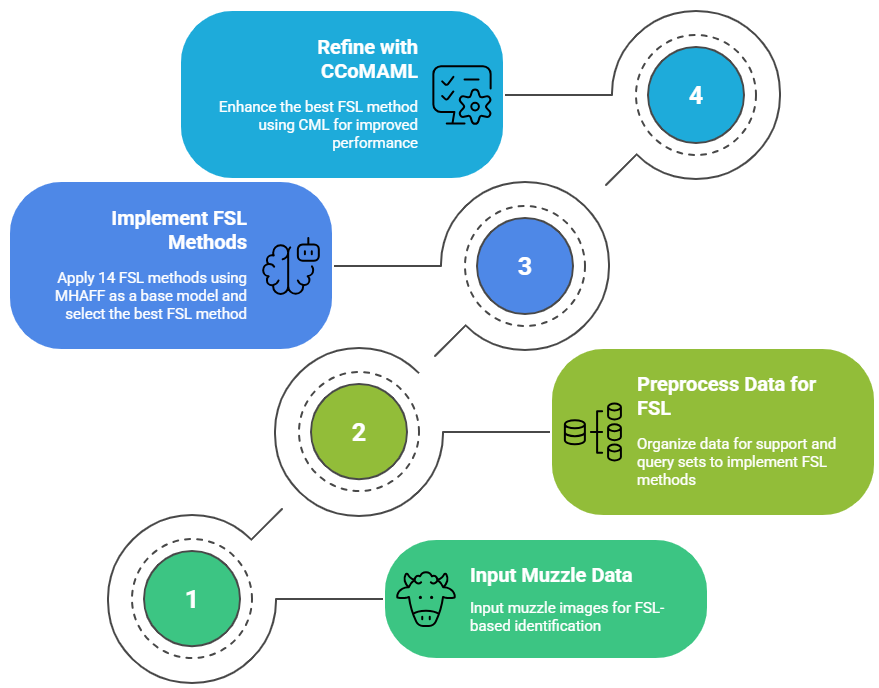}
    \caption{An overview of the methodology of this research.}
    \label{methodology_fig}
\end{figure}

Figure~\ref{methodology_fig} provides an overview of the research methodology. The process begins with the input data, which consists of muzzle images. In the context of few-shot learning, the base model serves as a foundational feature extractor that is trained on a broader dataset and later adapted to few-shot tasks. For this study, we use the MHAFF model~\citep{dulal2025mhaff} as the base model across all selected FSL methods due to its robust feature fusion capabilities and proven performance on small-scale datasets. MHAFF is a novel architecture that integrates features extracted by a CNN and a Vision Transformer (ViT) using an improved multi-head cross-attention mechanism. The model processes input images through two parallel branches: a CNN branch (Res-t) and a transformer branch (ViT-t), each extracting unique feature representations. These features are then fused using a multi-head attention mechanism. The fused representation is passed through a fully connected layer as a final feature representation. MHAFF outperformed CNN and transformer-based models, showcasing a promising base model~\citep{dulal2025mhaff}. The next step is the training of the selected fourteen FSL methods. Each method is rigorously tested to assess its effectiveness. Based on this evaluation, the best-performing FSL method is chosen to develop the proposed Cattle identification method using Cooperative Meta-Learning (CCoMAML). This section details the execution of the pipeline presented in Fig.~\ref{methodology_fig}.

\subsection{Datasets Description}
This section begins with an introduction to the datasets used in this research, followed by the steps taken to prepare them for FSL experiments. 

This study used two publicly available cattle datasets and two benchmark datasets. UNE muzzle~\citep{shojaeipour2021automated} and UNL muzzle data~\citep{li2022individual} are cattle datasets used in this research. UNE data was captured at the University of New England (UNE) farm. This data includes 2632 cattle face images of 300 cattle. The muzzles were detected and extracted from facial image data using a modified YOLOv5 model developed in our previous research~\citep{dulal2022automatic}. After removing blurred images from the extracted muzzle, the resulting muzzle dataset comprised a total of 2,447 images. UNL muzzle~\citep{li2022individual} data was captured at the University of Nebraska-Lincoln (UNL) Eastern Nebraska Research Extension and Education Center (ENREEC)'s farm. This data includes 4923 images from 268 cattle. The available images are muzzle regions obtained from manual cropping. 

\begin{figure}[ht]
\centering
\begin{tabular}{>{\raggedleft\arraybackslash}m{1.5cm} >{\centering\arraybackslash}m{2cm} >{\centering\arraybackslash}m{2cm} >{\centering\arraybackslash}m{2cm} >{\centering\arraybackslash}m{2cm} >{\centering\arraybackslash}m{2cm}}
 &  &  &  &  &  \\ 

\rotatebox{90}{CIFAR10} & \includegraphics[width=2cm,height=2cm]{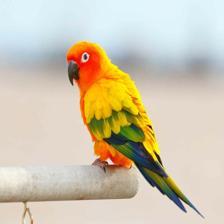} & \includegraphics[width=2cm,height=2cm]{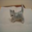} & \includegraphics[width=2cm,height=2cm]{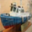} & \includegraphics[width=2cm,height=2cm]{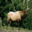} & 
\includegraphics[width=2cm,height=2cm]{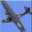}  \\ 

\rotatebox{90}{Flower102} & \includegraphics[width=2cm,height=2cm]{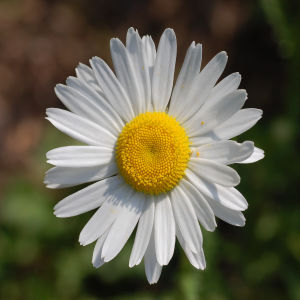} & \includegraphics[width=2cm,height=2cm]{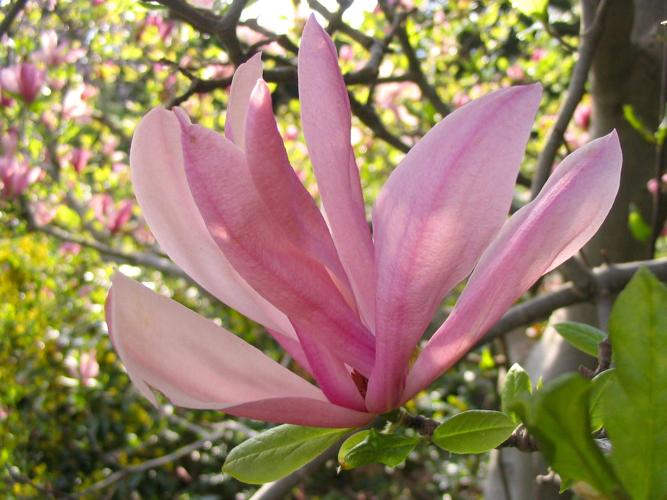} & \includegraphics[width=2cm,height=2cm]{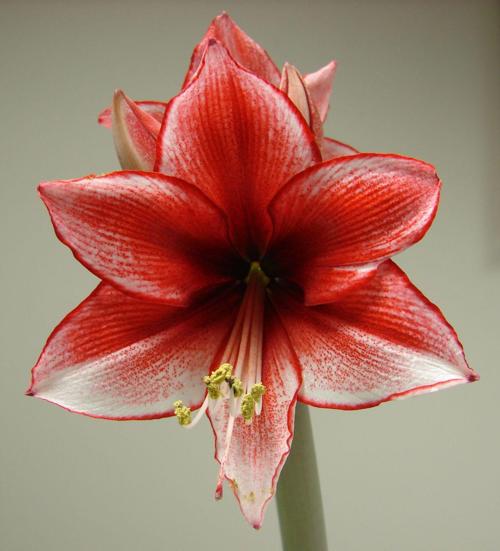} & \includegraphics[width=2cm,height=2cm]{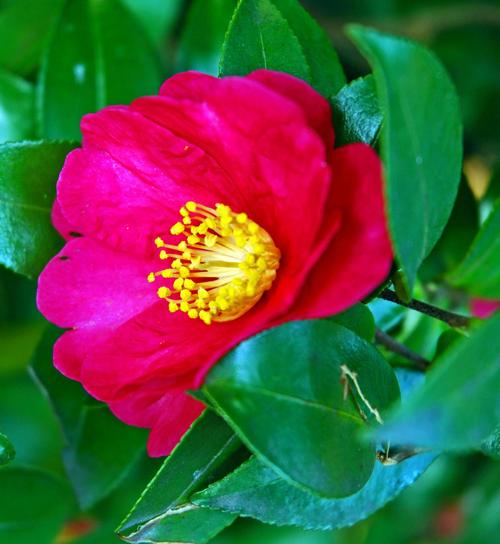} & \includegraphics[width=2cm,height=2cm]{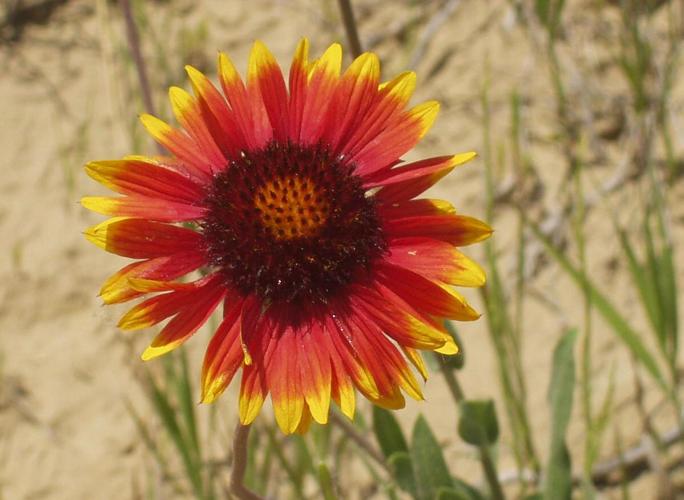}   \\ 

\rotatebox{90}{UNE muzzle} & \includegraphics[width=2cm,height=2cm]{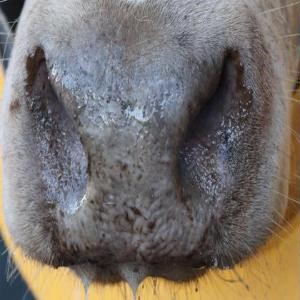} & \includegraphics[width=2cm,height=2cm]{UNE1.JPG} & \includegraphics[width=2cm,height=2cm]{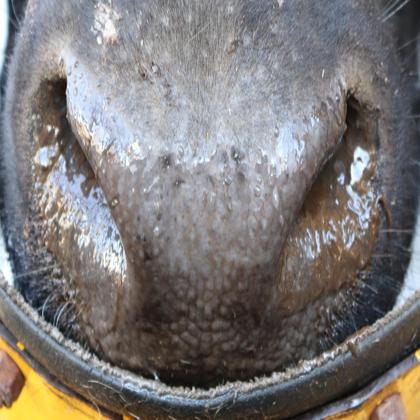} & \includegraphics[width=2cm,height=2cm]{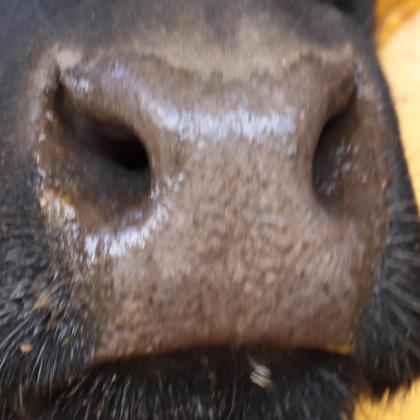} & \includegraphics[width=2cm,height=2cm]{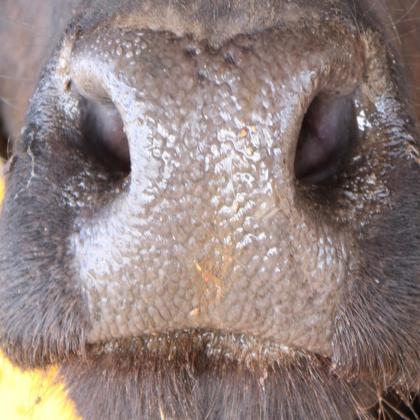}   \\ 

\rotatebox{90}{UNL muzzle} & \includegraphics[width=2cm,height=2cm]{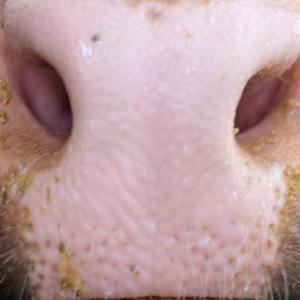} & \includegraphics[width=2cm,height=2cm]{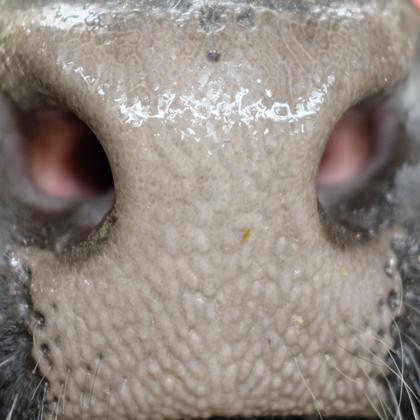} & \includegraphics[width=2cm,height=2cm]{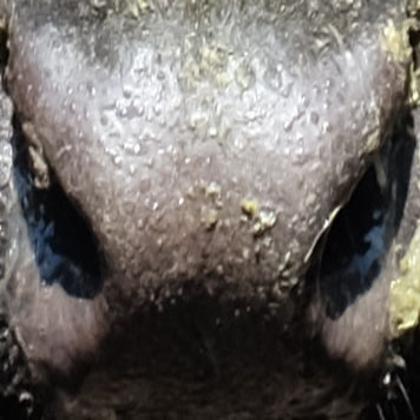} & \includegraphics[width=2cm,height=2cm]{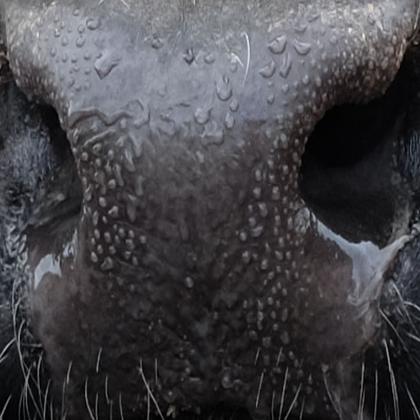} & \includegraphics[width=2cm,height=2cm]{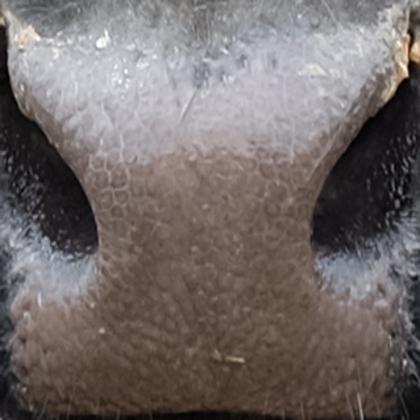}  \\ 

\end{tabular}
\caption{Representative samples from each dataset, showcasing sample images from different classes in each dataset.}
\label{sample_four_data}
\end{figure}

The benchmark datasets, CIFAR10~\citep{krizhevsky2009learning} and Flower102~\citep{nilsback2008automated}, were used to validate the CCoMAML due to their well-established use. CIFAR10 is a commonly used dataset for image classification. It has a total of 60000 images with ten classes. CIFAR10 is a useful benchmark for evaluating a model's ability to classify images across diverse object categories. It can help assess the general performance of the model in differentiating between distinct classes. Flower102, also known as the Oxford 102 Flowers dataset, is an image dataset designed for fine-grained visual categorization. It comprises 102 different classes of flowers. The total number of images is 8169. Flower102 is a fine-grained dataset like the cattle muzzle dataset. It involves distinguishing between classes with subtle differences in appearance. For example, different flower species may appear visually similar, just as different cattle muzzles can have similar features but slight differences. 
Fig.~\ref{sample_four_data} shows sample images of all four datasets.  

\subsubsection{Data preparation for FSL Methods}
\label{dataset_fsl}
The datasets in few-shot learning are typically divided into two subsets: the support set and the query set. The support set contains a few labeled examples (e.g. K per class) that the model uses to learn or adapt to new cattle classes. The query set consists of unlabeled muzzle images from the same N classes, which the model must classify after learning from the support set. This structure simulates real-world scenarios where a model encounters new, unseen data and must make accurate predictions based on limited prior information~\citep{hospedales2021meta}. Constructing the support and query sets requires careful selection to ensure that each class is adequately represented despite the limited number of examples. During training, multiple few-shot tasks are created by randomly selecting different classes and forming corresponding support and query sets. Such a strategy, called episodic training, helps the model learn how to learn, improving its ability to rapidly adapt to new tasks with minimal data~\citep{martin2022towards, wang2019hybrid, bennequin2021bridging, wang2023few}. FSL methods typically use disjoint class sets for training and testing, ensuring that the model learns a transferable initialization that can quickly adapt to new tasks with unseen classes~\citep{snell2017prototypical, triantafillou2019meta}.

In this experiment, the CIFAR10 dataset was split into 5 classes for training and the remaining 5 classes for testing. 
For the Flower102 dataset, 61 classes were used for training, and the remaining 41 classes were used for testing. For the cattle muzzle data, the UNE muzzle dataset is used for training, while the UNL muzzle dataset is used for testing. For all datasets, the image size was set to $224 \times 224$ following the base model's requirement.

In few-shot learning, the terms N-way and K-shot are commonly used to define the learning tasks. An N-way classification task involves N distinct classes that the model must differentiate between. A K-shot task provides K-labeled examples per class for training. For instance, a 5-way 1-shot task requires the model to classify among five cattle classes with only one labeled muzzle per class, while a 10-way 5-shot task involves ten classes with five muzzles each. This N-way K-shot framework is essential for evaluating and comparing few-shot learning models under standardized conditions~\citep{li2021meta}.

A \textbf{task} in FSL refers to a specific learning problem or mini-dataset sampled from a larger distribution. Each task is typically divided into two sets: the \textbf{support set} and the \textbf{query set}. The support set contains a small number of labeled examples from each class and serves as a reference for learning. It acts like a miniature training set that helps the model understand the characteristics of each class within the context of the current task. In contrast, the query set consists of unlabeled examples from the same set of classes, and the model's goal is to predict their labels based on what it learned from the support set. Essentially, the support set provides the "few shots" of information the model learns from, while the query set tests the model's ability to generalize from that limited information. This setup mimics the way humans can recognize new categories after seeing only a few examples.

\subsection{Selection of FSL Method}
The primary objective of this phase is to identify the most effective FSL method. The evaluation is conducted using MHAFF as the base model. A total of fourteen FSL methods are selected and systematically evaluated to assess their performance. The selected methods include Siamese Networks, Prototypical Networks, Matching Networks, Relation Networks, MAML, Reptile, SNAIL, Meta-SGD, FOMAML, MAML++, ANIL, BOIL, LEO, and CML. Siamese, Prototypical Networks, and MAML were included based on their widespread use in prior cattle identification research, as presented in the related work section. For a more comprehensive evaluation, additional FSL methods were selected, covering a diverse range of learning paradigms. Siamese, Prototypical, Matching, and Relation Networks represent metric-based approaches, which rely on learning similarity measures. In contrast, MAML, Reptile, SNAIL, Meta-SGD, FOMAML, MAML++, ANIL, BOIL, LEO, and CML fall under optimization-based methods that focus on learning adaptable initialization or update strategies. Including both types of methods enables a rigorous comparison.

The best-performing method is taken as the baseline, and further improvements are made to develop the proposed method in this research.

\subsection{Proposed Cattle Identification Method}
CCoMAML is a novel cattle identification method proposed in this research, built upon Cooperative MAML (CML)~\citep{shin2024cooperative}. Among the fourteen evaluated FSL methods, CML demonstrated superior performance and was therefore selected as the FSL strategy. Building upon CML, a custom co-learner module based on a CNN architecture is proposed, resulting in the complete CCoMAML framework.

\begin{figure}
    \centering
    \includegraphics[width=0.5\linewidth]{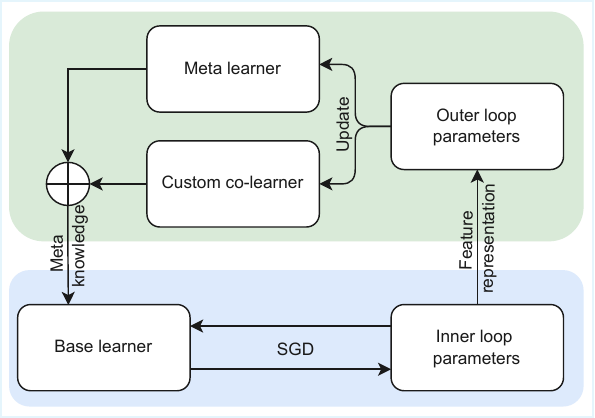}
    \caption{An overview of the CCoMAML method. The upper light green color represents the outer loop, and the lower light blue color represents the inner loop. The custom co-learner is a CNN-based architecture.}
    \label{coMAML}
\end{figure}

Cooperative MAML is an improved version of MAML that introduces a co-learner network to augment the gradient updates during meta-training, enhancing generalization across tasks without increasing inference complexity. An overview of the CML method is illustrated in Figure~\ref{coMAML}.

In this method, a meta-learner and a base-learner are the main players leading the learning process. The meta-learner captures shared knowledge across all learning tasks (e.g., identifying groups of cattle with different breeds at various locations) in a more general concept. The base learner focuses on learning task-specific features (identifying one group of cattle) using the initial knowledge provided by the meta-learner. 

During training, the base learner model updates its parameters using gradient descent, like Stochastic Gradient Descent (SGD)~\citep{bottou2012stochastic} or a variant like Adam optimizer~\citep{adam2014method}, for individual cattle identification tasks with the guidance of the meta-learner and computes the loss, which is then passed back to the meta-learner. Over multiple tasks, the meta-learner aggregates these losses and updates its parameters to refine the meta-knowledge to be shared. This process enables the base learner to quickly adapt and perform well on new cattle identification tasks with minimal training. MAML learns meta-knowledge in two loops: the inner loop and the outer loop. The inner loop focuses on task-specific learning to update the task-specific parameters, \(\phi\). The outer loop tries to optimize meta-knowledge, represented by ($\theta$). The outer loop parameters are generated across all tasks to obtain a general understanding to learn how to identify one group of cattle and allow rapid adaptation to another new group of cattle. Hence, the main objective of MAML during training is to find the generalized or the best parameters of the base learner
by updating through the process of inner and outer loops. In simple words, the outer loop primarily handles the extraction of generalizable features across tasks, establishing a robust foundation for learning. The inner loop then performs task-specific adaptations, fine-tuning these features to meet the requirements of individual tasks. This division of labor enables MAML to achieve rapid and effective learning from limited data~\citep{raghu2019rapid, starshak2022negative}.

The base learner is trained to solve individual tasks. During the inner loop, the base learner adapts its parameters using the support set of a specific task through a few gradient descent steps. Its role is to quickly learn task-specific knowledge with limited data. After this adaptation, the base learner is evaluated on the query set, and the resulting loss is used in the outer loop to update the initial parameters (meta-parameters). This helps the base learner become better at adapting to new tasks with minimal data. MAML is applied to optimize the parameters \(\theta\) of MHAFF such that the base learner can rapidly adapt to new cattle identification tasks using \(\theta\) with minimal training data. In such a way, MAML enables MHAFF to achieve robust generalization and efficient learning with few labeled muzzle images.

The meta-learner is a higher-level optimization process that works with the base learner's architecture (MHAFF) to optimize the meta-parameters (\(\theta\)) by training across multiple cattle identification tasks in one particular cattle muzzle dataset. This optimization occurs in the outer loop, where the meta-learner updates \(\theta\) based on the performance of the base learner in the inner loop, which performs task-specific adaptation. During testing, the learned \(\theta\) is initialized for fine-tuning MHAFF on new cattle identification tasks (different than the trained dataset) with minimal muzzle images.

Thus the equation~\ref{problem_formulation_equation} can be represented as:
\begin{equation}
f_{\theta}^{\text{MHAFF}}(\mathcal{D}_\mathcal{T}, \mathcal{L}_{\mathcal{D}_\mathcal{T}}) 
\rightarrow 
f_{\phi}^{\text{MHAFF}}(\mathcal{D}_{\mathcal{T}_j}, \mathcal{L}_{\mathcal{D}_{\mathcal{T}_j}}).
\end{equation}

\subsubsection{Optimization of Inner Loop Parameters}
The inner loop is responsible for adapting to a specific task. It begins with a shared set of meta-parameters \(\theta\) that initializes all tasks. Given a new task with support data $D^{Su}$, the model computes the task-specific loss using a small dataset associated with that task. This loss updates the model parameters through a few gradient descent steps. The updated parameters, denoted as \(\phi^{\prime}\), 
 are explicitly optimized for the given task. Support data is used in the inner loop to train and update the parameters based on the task-specific loss. The inner loop parameters are updated by using the equation~\ref{inner_loop}.
 
\begin{equation} \label{inner_loop}
\phi_i^{\prime} = \theta - \alpha \nabla_{\theta} \mathcal{L}(\theta, \mathcal{D}_i^{\text{Su}})
\end{equation}
where  
\( \theta \) represents the initial meta-parameters,  
\( \alpha \) represents the inner loop learning rate,  
\( \mathcal{L}(\theta, \mathcal{D}_i^{\text{Su}}) \) represents the loss function of the inner loop with Support data of that specific task,  
\( \nabla_{\theta} \) represents the gradient descent, and \(i \) represents the i-th task. 
After adapting the model to each task using the support set, the task-specific parameters (obtained in the inner loop) are used in the outer loop to evaluate performance on the corresponding query set and update the shared meta-parameters.

\subsubsection{Optimization of Outer Loop Parameters}
 The outer loop optimizes the initial meta-parameters \(\theta\) to lead to fast adaptation in the inner loop. After performing task-specific updates in the inner loop, the outer loop evaluates the updated inner loop parameters 
\(\phi^{\prime}\) on query data. The objective of the outer loop is to minimize the loss of meta-optimization using the outer loop loss calculation. This loss further optimizes the meta parameters to \(\theta^{\prime}\). The \(\theta^{\prime}\) is optimized by the use of the equation~\ref{meta_loop}.

\begin{equation} \label{meta_loop}
\theta_i^{\prime} = \theta - \beta \nabla_{\theta} \mathcal{L}_{meta}(\phi_i^{\prime}, \mathcal{D}_i^{\text{Qu}})
\end{equation}
where:    
\( \beta \) represents the outer loop learning rate,  
\(\mathcal{L}_{meta}(\phi_i^{\prime}, \mathcal{D}_i^{\text{Qu}})\) represents the meta loss in the outer loop with query data.

CCoMAML works by performing gradient modification in the outer loop by using a co-learner~\citep{shin2024cooperative}. Gradient modification refers to the deliberate alteration of gradients, typically by introducing structured or learnable noise, to influence how the model parameters are updated during meta-training. Rather than relying solely on raw gradients, this approach adds a layer of control that can guide the optimization trajectory, helping the model converge towards solutions that generalize better across tasks. This regularization enhances generalization by encouraging smoother gradient updates~\citep{shin2024cooperative, finn2018probabilistic}. During meta training, gradient noise is imposed in the outer loop as an augmented parameter gradient. The augmented parameter gradient is the original gradient and additional gradients, often introduced through auxiliary models or external transformations called a co-learner. It is an auxiliary model that operates alongside the primary meta-learner. Its primary function is to generate an alternative gradient by injecting learnable noise into the gradient computations. This process provides a form of regularization that encourages finding more generalizable meta-initialization parameters. 

The co-learner is not involved in the task-specific optimization in the inner loop. Its parameters are only updated in the outer loop using query data. The co-learner operates alongside the process of meta-optimization of parameters \(\theta_i^{\prime}\),  aided by the inner loop optimized parameters \(\phi_i^{\prime}\). The co-learner does not know the current task (inside the inner loop) but retains information about the previous task through \(\phi_{i-1}^{\prime}\). If \(\psi\) represents the co-learners' parameters, they are updated using the loss function as shown in Equation~\ref{co_learner_loop}.  

\begin{equation} \label{co_learner_loop}
\psi_i^{\prime} = \psi - \beta \nabla_{\psi} \mathcal{L}_{co}(\phi_{i-1}^{\prime}, \mathcal{D}_i^{\text{Qu}})
\end{equation}

where $\beta$ is the learning rate for the outer loop.

The main objective of the outer loop is to minimize the total loss, the outer loop combines the meta-loss and the co-learner loss using a controlled factor \(\gamma\), where \(\gamma \in [0, 1]\) represents the level of noise introduced in the outer loop via the co-learner. Additionally, a weight decay of \(1 \times 10^{-5}\) is applied to the final loss as an L2 regularization term (\(\mathcal{R}_{\text{L2}}\)). The total loss \(\mathcal{L}_{total}\) is computed using Equation~\ref{total_loss}.  

\begin{equation} \label{total_loss}
\mathcal{L}_{total} = \mathcal{L}_{meta} + \gamma\ \mathcal{L}_{co} + \mathcal{R}_{\text{L2}}
\end{equation}

Since both meta-parameters and co-learner parameters are updated in the outer loop, the final parameters of the outer loop (\(\theta^{\prime}\) and \(\psi^{\prime}\)) are optimized together using the total loss calculated on the query data. Following the findings from the MHAFF study~\citep{dulal2025mhaff}, cross-entropy loss~\citep{mao2023cross} demonstrated superior performance. Therefore, cross-entropy loss is adopted in this research.

\begin{figure}
    \centering
    \includegraphics[width=0.8\linewidth]{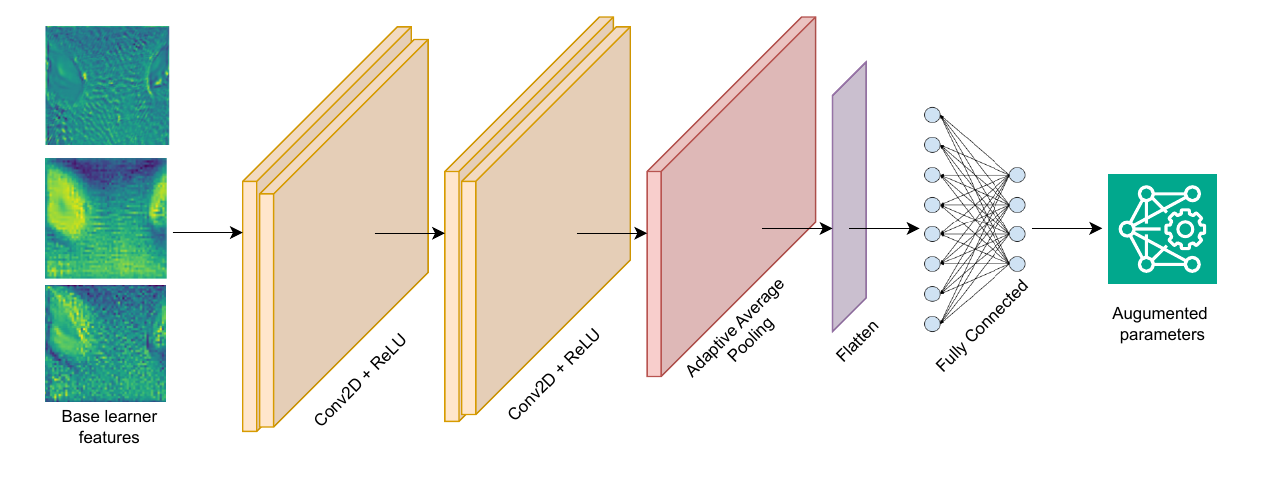}
    \caption{Architecture of the proposed custom CNN-based co-learner module. The co-learner takes feature representations as input and leverages the outer loop’s initial parameters to generate augmented gradients or updated parameters through a series of convolutional and fully connected layers.}
    \label{custom_colearner}
\end{figure}

Figure~\ref{custom_colearner} illustrates the architecture of the proposed custom co-learner, which adopts a simple CNN design. It consists of two convolutional layers with ReLU activation, followed by an adaptive average pooling layer and two fully connected layers. A comprehensive explanation of its design rationale and development process is provided in Section~\ref{fsl_abalation_studies}.

During the adaptation to a new task in testing, the inner loop fine-tunes the MHAFF model on new, unseen tasks using the learned meta-parameters \(\theta^{\prime}\). The model's performance is then evaluated on the corresponding query data. The support sets are used for computing the task-specific gradients and updating the model parameters for each task individually using a few gradient steps.
While the query sets are used to evaluate the adapted model's performance on the new task, measuring how well the meta-learned initialization generalizes to unseen data. 

\subsection{Evaluation Metrics}
The model's performance was evaluated using two key metrics: Accuracy and F1 Score. Accuracy represents the proportion of correct predictions, while F1 Score balances false positives and false negatives. These metrics are computed as follows:

\begin{align}
    \text{Accuracy} &= \frac{TP + TN}{TP + TN + FP + FN} \\
    \text{F1 Score} &= \frac{2TP}{2TP + FP + FN}
\end{align}

where, \( TP \) (true positives) and \( TN \) (true negatives) are correctly predicted positive and negative samples, respectively; \( FP \) (false positives) are incorrectly predicted positives, and \( FN \) (false negatives) are actual positives incorrectly predicted as negatives.

\section{Experiments and Results}
\label{experiments and results}
This section presents the training details, experimental results, and corresponding analyses conducted in this study. All experiments were conducted on a system equipped with a Gold 5315Y processor and an NVIDIA RTX 3080 GPU with 64GB of memory. The implementation was carried out using Python 3.13 and built on the PyTorch framework. All selected few-shot learning (FSL) methods were trained for 200 epochs. To prevent overfitting, several optimization strategies were employed, including dynamic learning rate adjustment and early stopping. The Adam optimizer was used in combination with the ReduceLROnPlateau scheduler to dynamically adjust the learning rate. As in MHAFF~\citep{dulal2025mhaff}, MHAFF was implemented using PyTorch's library with transfer learning of the pre-trained weights from the ImageNet~\citep{deng2009imagenet} dataset.
The training process employs specific hyperparameters to ensure efficient optimization and generalization. The inner-loop learning rate is set to \(1 \times 10^{-1}\), enabling fast updates for task-specific learning. In comparison, the meta-learning rate is set to \(1 \times 10^{-3}\) to allow for stable, gradual improvements in the model’s generalization across tasks. The model and the co-learner component are optimized using the Adam optimizer, known for its adaptive learning rate, which helps accelerate convergence while maintaining stability. A weight decay of \(1 \times 10^{-5}\) is applied to both optimizers, acting as an L2 regularization technique to prevent overfitting. These carefully selected values promote effective meta-learning, enabling the model to learn efficiently from each task while generalizing well across diverse tasks. The initial learning rates were reduced by a factor of 0.1 if the validation loss did not improve for 10 consecutive epochs. Early stopping was triggered if there was no improvement in validation loss for 20 epochs. 

The experiments in this section are organized into four parts. Section~\ref{fsl_perf_comp} presents the performance comparison of 14 few-shot learning (FSL) methods, and compares the results with the proposed approach. Section~\ref{fsl_abalation_studies} details the ablation studies conducted to develop the proposed CCoMAML method. Section~\ref{fsl_cross_validation} provides cross-validation results obtained by swapping the training and testing splits of the muzzle datasets, offering insights into the model’s generalization capability. Section~\ref{DLvsFSL} presents a comprehensive evaluation of DL models with CCoMAML under constrained data conditions, where the number of training samples per class varies from 1 to 9. This experiment is designed to systematically examine the performance degradation of conventional deep learning approaches with limited annotated data and to provide a direct and fair comparison with the few-shot learning framework, CCoMAML, evaluated under identical conditions.

\subsection{Performance Comparison With Other FSL Methods}
\label{fsl_perf_comp}
This section presents the results used to identify the most effective FSL method among the fourteen comprehensively evaluated approaches. In addition, it includes a comparative analysis of the proposed CCoMAML method against these 14 FSL methods. The evaluation is conducted under standard 5-way 5-shot and 5-way 1-shot settings, using accuracy and F1 score as performance metrics. These configurations are widely adopted benchmarks in FSL research, enabling consistent and fair comparisons across different methods, where the 5-way 5-shot setting is considered relatively easier due to more examples per class, and the 5-way 1-shot setting poses a greater challenge due to limited examples. The comparative results are presented in Table~\ref{fsl_results_comparecifar} for CIFAR-10, Table~\ref{fsl_results_comparisonflower} for Flower102, and Table~\ref{fsl_results_comparison} for the cattle muzzle dataset. To provide a more reliable estimate of model performance, accuracy is reported along with a 95\% confidence interval (CI), which indicates the range within which the true accuracy is expected to lie with 95\% certainty. This accounts for variability in the data and offers a better understanding of the model's generalization, especially in limited data settings.

\begin{table}[ht]
\centering
\footnotesize
\setlength{\tabcolsep}{2pt}
\caption{Comparison of the accuracy and F1 score (mean percentage $\pm$ 95\% CI) of few-shot learning algorithms on CIFAR10. Results are averaged over 5 runs.}
\label{fsl_results_comparecifar}
\begin{tabular}{@{}l l c c c c@{}}
\toprule
\makecell{\textbf{Learning} \\ \textbf{Strategy}} & \textbf{Method} & \multicolumn{2}{c}{\textbf{5-way 1-shot}} & \multicolumn{2}{c}{\textbf{5-way 5-shot}} \\
\cmidrule(lr){3-4} \cmidrule(lr){5-6}
& & Accuracy & F1 & Accuracy & F1 \\
\midrule
\multirow{4}{*}{Metric}
    & Siamese            & $58.32 \pm 2.38$ & $60.10 \pm 2.13$ & $64.35 \pm 2.85$ & $65.10 \pm 2.17$ \\
    & Protonet           & $59.50 \pm 2.25$ & $61.00 \pm 1.94$ & $66.12 \pm 2.80$ & $67.15 \pm 2.12$ \\
    & Matching Networks  & $57.65 \pm 2.61$ & $58.23 \pm 2.25$ & $62.54 \pm 2.91$ & $63.40 \pm 2.28$ \\
    & Relation Networks  & $57.22 \pm 2.42$ & $59.40 \pm 2.18$ & $63.25 \pm 2.95$ & $64.85 \pm 2.21$ \\
\midrule
\multirow{11}{*}{Optimization}
    & MAML               & $60.03 \pm 2.30$ & $61.65 \pm 2.00$ & $66.89 \pm 2.64$ & $67.62 \pm 2.08$ \\
    & Reptile            & $57.20 \pm 2.58$ & $57.95 \pm 2.29$ & $64.21 \pm 2.83$ & $65.02 \pm 2.13$ \\
    & SNAIL              & $57.10 \pm 2.45$ & $58.72 \pm 2.12$ & $64.42 \pm 2.81$ & $65.21 \pm 2.24$ \\
    & Meta-SGD           & $58.30 \pm 2.35$ & $60.10 \pm 2.14$ & $64.80 \pm 2.71$ & $66.73 \pm 2.09$ \\
    & FOMAML             & $58.12 \pm 2.48$ & $58.94 \pm 2.20$ & $63.95 \pm 2.87$ & $64.99 \pm 2.16$ \\
    & MAML++             & $59.92 \pm 2.39$ & $60.21 \pm 2.03$ & $65.91 \pm 2.69$ & $66.21 \pm 2.05$ \\
    & ANIL               & $58.75 \pm 2.46$ & $60.02 \pm 2.18$ & $64.98 \pm 2.74$ & $66.85 \pm 2.08$ \\
    & BOIL               & $58.95 \pm 2.53$ & $60.30 \pm 2.13$ & $65.40 \pm 2.69$ & $66.20 \pm 2.14$ \\
    & LEO                & $61.00 \pm 2.36$ & $60.85 \pm 2.01$ & $66.42 \pm 2.63$ & $67.60 \pm 2.00$ \\
    & CML                & $63.65 \pm 2.30$ & $64.88 \pm 1.34$ & $68.22 \pm 2.61$ & $68.70 \pm 2.01$ \\
    & \textbf{CCoMAML}   & $\mathbf{66.05 \pm 3.78}$ & $\mathbf{68.55 \pm 0.83}$ & $\mathbf{70.89 \pm 2.41}$ & $\mathbf{71.11 \pm 1.38}$ \\
\bottomrule
\end{tabular}
\end{table}

\begin{table*}[ht]
\centering
\footnotesize
\setlength{\tabcolsep}{2pt}
\caption{Comparison of the accuracy and F1 score (mean percentage $\pm$ 95\% CI) of few-shot learning algorithms on Flower102. Results are averaged over 5 runs.}
\label{fsl_results_comparisonflower}
\begin{tabular}{@{}l l c c c c@{}}
\toprule
\multirow{2}{*}{\makecell{Learning \\ Strategy}} & Method & \multicolumn{2}{c}{5-way 1-shot} & \multicolumn{2}{c}{5-way 5-shot} \\
\cmidrule(lr){3-4} \cmidrule(lr){5-6}
& & Accuracy & F1 & Accuracy & F1 \\
\midrule
\multirow{4}{*}{Metric}
& Siamese            & $60.72 \pm 2.85$ & $59.91 \pm 3.11$ & $75.42 \pm 2.61$ & $72.98 \pm 2.85$ \\
& Protonet           & $63.12 \pm 3.04$ & $63.58 \pm 3.21$ & $76.85 \pm 2.72$ & $76.18 \pm 2.93$ \\
& Matching Networks  & $60.03 \pm 3.21$ & $58.04 \pm 3.41$ & $70.57 \pm 3.05$ & $69.87 \pm 3.25$ \\
& Relation Networks  & $60.78 \pm 3.12$ & $60.43 \pm 3.34$ & $72.01 \pm 2.91$ & $72.46 \pm 3.14$ \\
\midrule
\multirow{11}{*}{Optimization}
& MAML              & $66.31 \pm 2.71$ & $65.75 \pm 2.92$ & $78.92 \pm 2.52$ & $77.39 \pm 2.69$ \\
& Reptile           & $61.28 \pm 3.09$ & $58.88 \pm 3.19$ & $74.49 \pm 2.79$ & $74.27 \pm 3.02$ \\
& SNAIL             & $61.67 \pm 2.94$ & $60.78 \pm 3.11$ & $76.18 \pm 2.59$ & $73.48 \pm 2.69$ \\
& Meta-SGD          & $63.29 \pm 2.84$ & $63.91 \pm 3.05$ & $76.54 \pm 2.62$ & $76.32 \pm 2.79$ \\
& FOMAML            & $63.19 \pm 3.01$ & $62.59 \pm 3.23$ & $74.68 \pm 2.73$ & $74.17 \pm 2.88$ \\
& MAML++            & $65.72 \pm 2.84$ & $63.29 \pm 2.91$ & $77.32 \pm 2.53$ & $76.97 \pm 2.63$ \\
& ANIL              & $65.39 \pm 2.71$ & $64.62 \pm 2.83$ & $77.86 \pm 2.42$ & $76.57 \pm 2.54$ \\
& BOIL              & $63.76 \pm 2.81$ & $64.02 \pm 2.93$ & $77.12 \pm 2.54$ & $75.49 \pm 2.62$ \\
& LEO               & $68.03 \pm 2.58$ & $66.38 \pm 2.75$ & $78.39 \pm 2.45$ & $76.82 \pm 2.56$ \\
& CML               & $67.56 \pm 2.42$ & $66.41 \pm 2.63$ & $82.66 \pm 2.30$ & $82.97 \pm 2.39$ \\
& \textbf{CCoMAML}  & $\mathbf{70.10 \pm 2.31}$ & $\mathbf{70.27 \pm 2.54}$ & $\mathbf{85.73 \pm 2.21}$ & $\mathbf{85.38 \pm 2.34}$ \\
\bottomrule
\end{tabular}
\end{table*}

\begin{table}[ht]
\centering
\footnotesize
\setlength{\tabcolsep}{2pt}
\caption{Comparison of the accuracy and F1 score (mean percentage $\pm$ 95\% CI) of few-shot learning algorithms on muzzle data. Results are averaged over 5 runs. An asterisk (*) indicates statistical significance based on a paired \textit{t}-test ($p < 0.05$), and double asterisks (**) indicate high significance ($p < 0.001$) when compared with CCoMAML.}
\label{fsl_results_comparison}
\begin{tabular}{@{}l l c c c c@{}}
\toprule
\makecell{\textbf{Learning} \\ \textbf{Strategy}} & \textbf{Method} & \multicolumn{2}{c}{\textbf{5-way 1-shot}} & \multicolumn{2}{c}{\textbf{5-way 5-shot}} \\
\cmidrule(lr){3-4} \cmidrule(lr){5-6}
& & Accuracy & F1 & Accuracy & F1 \\
\midrule
\multirow{4}{*}{Metric} 
    & Siamese            & 83.55 $\pm$ 0.58$^*$ & 85.02 $\pm$ 0.57$^*$ & 89.65 $\pm$ 0.50$^{**}$ & 90.20 $\pm$ 0.48$^{**}$ \\
    & Protonet           & 85.12 $\pm$ 0.54$^*$ & 85.41 $\pm$ 0.52       & 92.88 $\pm$ 0.46$^*$ & 91.46 $\pm$ 0.44$^{**}$ \\
    & Matching Networks  & 81.75 $\pm$ 0.70$^*$ & 82.34 $\pm$ 0.65$^*$ & 88.65 $\pm$ 0.53$^{**}$ & 88.80 $\pm$ 0.54$^{**}$ \\
    & Relation Networks  & 82.92 $\pm$ 0.60$^*$ & 83.31 $\pm$ 0.61$^*$ & 89.72 $\pm$ 0.48$^{**}$ & 90.15 $\pm$ 0.42$^{**}$ \\
\midrule
\multirow{11}{*}{Optimization}
    & MAML               & 85.71 $\pm$ 0.58$^*$ & 85.78 $\pm$ 0.54$^*$ & 93.42 $\pm$ 0.47$^*$ & 92.88 $\pm$ 0.44$^{**}$ \\
    & Reptile            & 81.98 $\pm$ 0.69$^*$ & 81.80 $\pm$ 0.68$^*$ & 91.06 $\pm$ 0.53$^*$ & 91.48 $\pm$ 0.50$^{**}$ \\
    & SNAIL              & 82.49 $\pm$ 0.64$^*$ & 82.95 $\pm$ 0.62$^*$ & 89.96 $\pm$ 0.50$^{**}$ & 90.03 $\pm$ 0.45$^{**}$ \\
    & Meta-SGD           & 83.62 $\pm$ 0.58$^*$ & 83.98 $\pm$ 0.57$^*$ & 91.46 $\pm$ 0.48$^{**}$ & 91.98 $\pm$ 0.42$^{**}$ \\
    & FOMAML             & 82.73 $\pm$ 0.62$^*$ & 83.10 $\pm$ 0.60$^*$ & 90.67 $\pm$ 0.52$^{**}$ & 91.01 $\pm$ 0.49$^{**}$ \\
    & MAML++             & 83.94 $\pm$ 0.57$^*$ & 84.05 $\pm$ 0.54$^*$ & 91.88 $\pm$ 0.44$^{**}$ & 92.35 $\pm$ 0.40$^{**}$ \\
    & ANIL               & 83.62 $\pm$ 0.62$^*$ & 84.21 $\pm$ 0.58$^*$ & 91.89 $\pm$ 0.45$^{**}$ & 92.22 $\pm$ 0.39$^{**}$ \\
    & BOIL               & 83.45 $\pm$ 0.66$^*$ & 83.85 $\pm$ 0.62$^*$ & 91.56 $\pm$ 0.43$^{**}$ & 91.88 $\pm$ 0.37$^{**}$ \\
    & LEO                & 84.22 $\pm$ 0.53$^*$ & 83.50 $\pm$ 0.54$^*$ & 91.40 $\pm$ 0.42$^{**}$ & 91.89 $\pm$ 0.35$^{**}$ \\
    & CML                & 85.40 $\pm$ 0.48$^*$ & 85.62 $\pm$ 0.50$^*$ & 94.88 $\pm$ 0.52$^*$ & 93.48 $\pm$ 0.49$^*$ \\
    & \textbf{CCoMAML}   & $ \mathbf{87.32} \pm \mathbf{0.31} $ & $ \mathbf{87.46} \pm \mathbf{0.29} $ & $ \mathbf{96.07} \pm \mathbf{0.28} $ & $ \mathbf{97.03} \pm \mathbf{0.25} $ \\
\bottomrule
\end{tabular}   
\end{table}

The results presented in Tables~\ref{fsl_results_comparecifar}, ~\ref{fsl_results_comparisonflower}, and ~\ref{fsl_results_comparison} demonstrate that CML is the best-performing method across all datasets in terms of both accuracy and F1 score in all settings. These results led us to select CML as the baseline FSL method for further development. Building on this, we enhance CML by introducing a novel co-learner module in the outer-loop optimization process, resulting in the proposed CCoMAML method.

CCoMAML achieves the highest performance across both 5-way 1-shot and 5-way 5-shot configurations. The superior results obtained on the Flower102 and cattle muzzle datasets, compared to CIFAR10, can be attributed to greater domain consistency and visual similarity among classes. Both Flower102 and the cattle muzzle dataset are fine-grained, with all classes (e.g., flower species or individual cattle) sharing similar structures and visual patterns, which facilitates effective knowledge transfer during meta-learning. In contrast, CIFAR10 comprises semantically diverse classes (e.g., airplanes vs. cats), resulting in higher inter-class variability and greater domain shift between training and testing classes—factors that hinder generalization~\citep{triantafillou2019meta, guo2020broader}.

The results in Table~\ref{fsl_results_comparison} demonstrate a clear superiority of CCoMAML, achieving the highest accuracy of $87.32 \pm 0.31\% $ for the 1-shot and $96.07 \pm 0.28\%$ for the 5-shot identification setting. This indicates a more robust ability to generalize from minimal data (as in the 1-shot setting) and exceptional scalability and adaptation with minimal additional support samples (as seen in the 5-shot case). Compared to traditional metric-based models such as Protonet ($85.12 \pm 0.54\%$ for 1-shot, $92.88 \pm 0.46$ for 5-shot) and optimization-based baselines like MAML ($85.71 \pm 0.58\%$ and $93.42 \pm 0.47\%$), CCoMAML outperforms them with a considerable margin of 1.34\% and 2.17\% in the respective tasks. Notably, the second-best method in the 5-shot setting is CML with $94.88 \pm 0.52\%$, still more than 1.1\% below CCoMAML, which significantly improves few-shot benchmarks. These results imply that CCoMAML effectively leverages cooperative and gradient-enhanced meta-learning to learn more transferable representations and adaptable decision boundaries. The lower standard deviation also underscores its stability and reliability across different runs, a critical factor in few-shot learning where variance can be high due to limited training data.

The F1 score, which balances precision and recall, offers additional insights into how well the models perform in situations with imbalanced or ambiguous class boundaries. Once again, CCoMAML performs best, obtaining $87.46 \pm 0.29\%$ in the 1-shot and a remarkable $97.03 \pm 0.25\%$ in the 5-shot setting. The improvement in F1 score over competing methods is particularly striking in the 5-shot case, where the margin over the next-best method (CML, at $93.48 \pm 0.49\%$) is more than three percentage, suggesting that CCoMAML is not only more accurate, but also significantly better at maintaining class balance and minimizing false positives or negatives. These findings suggest that CCoMAML produces more calibrated and confident predictions, even with sparse data, making it highly suitable for real-world applications where class imbalance or rare events are prevalent. The reduced standard deviation further indicates its robustness and consistency. Moreover, CCoMAML consistently reports the lowest standard deviation across all settings, highlighting its stability and reliability over three runs. 

To evaluate the statistical significance of performance differences between CCoMAML and other methods presented in Table~\ref{fsl_results_comparison}, a paired \textit{t}-test was performed. Overall, CCoMAML consistently outperformed competing methods with statistically significant improvements across most metrics, as shown by the resulting \textit{P}-values, except for the 5-way 1-shot F1 score compared to Protonet, where the difference was not significant. These results indicate that the observed performance gains of CCoMAML are unlikely to have occurred by chance. The strong statistical significance reflects CCoMAML’s more robust and reliable learning capability, likely due to its improved optimization strategies that enhance generalization in few-shot scenarios.

The metric-based FSL algorithms, such as Siamese Networks, Prototypical Networks, Matching Networks, and Relation Networks, rely on learning embedding spaces and computing similarity measures between support and query samples. While effective for modeling distance-based relationships, these approaches lack the flexibility of parameter adaptation during meta-training, limiting their ability to cope with high intra-class variability. On the other hand, optimization-based methods—including MAML, Reptile, SNAIL, Meta-SGD, FOMAML, MAML++, ANIL, BOIL, and LEO aim to learn initial parameters or adaptation strategies that quickly generalize to new tasks via gradient-based updates. However, these methods often face difficulty converging to sharp minima that do not transfer well to unseen distributions. CCoMAML addresses these limitations by introducing controlled stochasticity into the meta-gradient updates, enabling the model to escape poor local minima and regularize the learning process~\citep{shin2024cooperative}. Unlike CML, which also incorporates noise, the superior performance of CCoMAML over CML can be attributed to a key architectural advancement in noise generation. Architecturally, CCoMAML incorporates two convolutional layers followed by ReLU activations, an adaptive average pooling layer, and two fully connected layers that project features into a compact 64-dimensional noise vector. This structured pipeline preserves hierarchical representations and ensures fixed-dimensional, resolution-invariant outputs well-suited for noise-based gradient regularization. 

In contrast, CML also generates and injects noise during the meta-optimization of the outer loop in MAML but lacks the pooling layer and relies on directly flattening high-dimensional feature maps. This results in less controlled and potentially task-misaligned noise generation. Adaptive pooling and dedicated dimensionality reduction in CCoMAML allow it to produce a well-regularized and expressive noise vector that can be consistently integrated into the meta-learner’s gradient update, leading to more effective parameter augmentation across diverse tasks. Consequently, this design improves generalization on unseen tasks with minimal fine-tuning, as evidenced by its superior empirical performance. 

The enhanced performance of the proposed model is primarily due to the incorporation of a co-learner module. Co-learner adds learnable noise to the model's gradients during meta-training, effectively regularizing the learning process and improving generalization~\citep{shin2024cooperative, hinton2002stochastic, srivastava2014dropout, neelakantan2015adding}. Sharing the feature extractor with the base learner, the co-learner influences the same learned representations, ensuring coherent gradient updates. It provides efficient gradient augmentation without requiring additional forward passes or sub-network creation. By injecting controlled variability into the gradients, the co-learner prevents overfitting to the limited training samples, enabling the base learner to discover meta-initialization parameters that adapt better to new tasks. This targeted gradient augmentation strengthens the model's generalization ability under few-shot learning conditions. This collective strategy allows the model to identify optimal meta-initialization parameters, enabling swift adaptation to new tasks with minimal data~\citep{shin2024cooperative}. Notably, the co-learner is only active during meta-training and is removed during meta-testing, ensuring no additional computational burden or performance loss during inference~\citep{shin2024cooperative}. 

\subsection{Abalation Study}
\label{fsl_abalation_studies}
This section presents a detailed overview of the co-learner module architecture and examines how varying the intensity of the co-learner affects the overall performance of cattle identification.

\subsubsection{Co-learner architecture}
A simple CNN was used for the co-learner due to its efficiency and low parameter count. Inspired by CML, we explored fully connected (FC) layers, adaptive average pooling (AAP), and convolutional (Conv) layers with ReLU. The design followed a systematic process, first selecting the number of FC layers, then testing various configurations as presented in Table~\ref{co-learner_strategy}, and finally determining the optimal number of convolutional layers.

\begin{table}[ht]
\centering
\footnotesize
\setlength{\tabcolsep}{6pt}
\caption{Design strategies for the co-learner module with component inclusion.}
\begin{tabular}{lccc}
\hline
Strategy & FC & AAP & Conv \\
\hline
S1 & \cmark & \xmark & \xmark \\
S2 & \cmark & \cmark & \xmark \\
S3 & \cmark & \xmark & \cmark \\
S4 & \cmark & \cmark & \cmark \\
\hline
\end{tabular}
\label{co-learner_strategy}
\end{table}

The accuracies of 5-way 1-shot and 5-shot experiments were evaluated to determine the optimal number of FC layers. The results are presented in Table~\ref{fc_co-learner}. The results indicate that increasing the number of FC layers initially improves the classification accuracy. Specifically, using two FC layers yields the best performance in both 1-shot (86.20\%) and 5-shot (93.98\%) tasks. While the 3-layer model achieves slightly lower 1-shot accuracy (86.10\%), its 5-shot accuracy (93.44\%) is lower compared to the 2-layer configuration. Interestingly, adding a fourth layer does not result in further improvement and leads to marginal performance drops in 1-shot accuracy. These findings suggest that using two FC layers provides an effective balance between model complexity and generalization, making it the most suitable configuration for few-shot learning in this context.

\begin{table}[ht]
\centering
\caption{Effect of the number of fully connected (FC) layers on 1-shot and 5-shot classification accuracy (\%).}
\begin{tabular}{ccc}
\hline
\textbf{FC layers} & \textbf{1-shot} & \textbf{5-shot} \\
\hline
1 & 85.70 & 91.932 \\
\textbf{2} & \textbf{86.20} & \textbf{93.98} \\
3 & 86.10 & 93.44 \\
4 & 85.93 & 93.98 \\
\hline
\end{tabular}
\label{fc_co-learner}
\end{table}

Following the number of FC layers, we further explored the configuration as presented in Table~\ref{co-learner_strategy}. Following the configuration with two FC layers, we further investigated a series of co-learner strategies to enhance performance, as presented in Table~\ref{co-learner-accuracy}. Strategy S1 represents the baseline configuration with two FC layers. In S2, an adaptive average pooling layer was added after the feature extraction stage, followed by the same two FC layers. S3 extends S1 by incorporating an additional convolutional layer before the FC layers. Finally, S4 combines the modifications from both S2 and S3—employing adaptive average pooling, a convolutional layer, and two FC layers. Among all configurations, S4 achieved the highest classification accuracy in both 1-shot (86.80\%) and 5-shot (95.31\%) settings, suggesting that the combination of spatial pooling and convolutional enhancement is beneficial for improving few-shot learning performance.

\begin{table}[ht]
\centering
\caption{Comparison of co-learner strategies using classification accuracy (\%) in 5-way 1-shot and 5-way 5-shot settings.}
\begin{tabular}{lcc}
\hline
\textbf{Strategy} & \textbf{1-shot} & \textbf{5-shot} \\
\hline
S1 & 86.00 & 93.98 \\
S2 & 86.10 & 94.04 \\
S3 & 86.49 & 94.88 \\
\textbf{S4} & \textbf{86.80} & \textbf{95.31} \\
\hline
\end{tabular}
\label{co-learner-accuracy}
\end{table}

Furthermore, building upon the S4 strategy, we conducted an additional experiment to investigate the impact of the number of convolutional layers. Specifically, we incrementally added convolutional layers to the architecture to evaluate their influence on classification performance. To further optimize the S4 strategy, we evaluated the effect of adding multiple convolutional layers prior to the fully connected components. As shown in Table~\ref{conv_layer_analysis}, introducing two convolutional layers yielded the highest accuracy in both 1-shot (87.32\%) and 5-shot (96.07\%) settings. Although increasing the number of convolutional layers beyond two did not improve performance. These results suggest that two convolutional layers offer an optimal balance between feature enrichment and model complexity in the co-learner configuration.

\begin{table}[ht]
\centering
\caption{Effect of the number of convolutional layers on 1-shot and 5-shot classification accuracy (\%) based on the S4 co-learner configuration.}
\begin{tabular}{ccc}
\hline
\textbf{Conv layers} & \textbf{1-shot} & \textbf{5-shot} \\
\hline
1 & 86.80 & 95.31 \\
2 & \textbf{87.32} & \textbf{96.07} \\
3 & 87.19 & 95.60 \\
4 & 87.21 & 95.49 \\
\hline
\end{tabular}
\label{conv_layer_analysis}
\end{table}

This improved performance is because of improved gradient quality and generalization ability. The use of adaptive pooling plays a critical role in standardizing feature representations, enabling the co-learner to produce consistent and meaningful gradient updates across tasks with varying spatial representations. Furthermore, the additional fully connected layer at the end enhances the model’s capacity to capture more complex feature interactions. As a result, this structure supports better gradient modulation in the outer loop, leading to improved meta-initialization and generalization across diverse tasks. 

\subsubsection{Effect of co-learner intensity in the proposed model}
\begin{figure}[ht]
\centering
\includegraphics[width=0.98\textwidth]{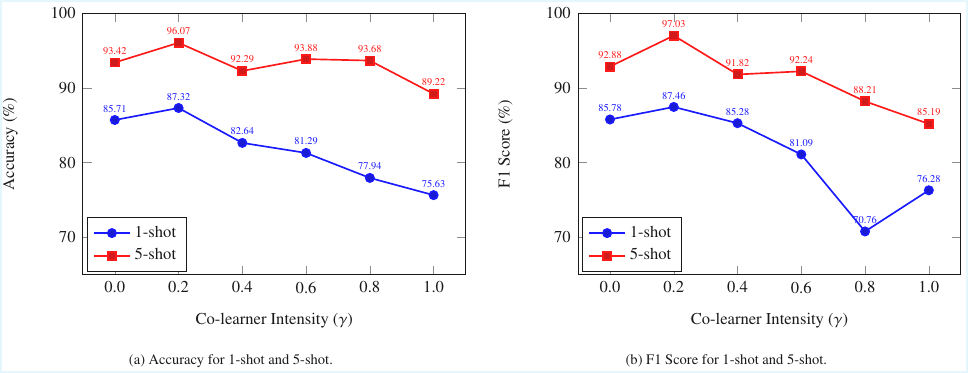}
\caption{Effect of co-learner intensity ($\gamma$) on CCoMAML's performance in 1-shot and 5-shot settings.}
\label{gamma_perf}
\end{figure}

This ablation experiment aims to find the best value of co-learner intensity in the proposed model. By introducing controlled noise into the gradients, the co-learner aids in discovering meta-initialization parameters that are more robust and capable of rapid adaptation across diverse tasks. This leads to a more effective starting point for fine-tuning new tasks~\citep{shin2024cooperative}. The value of $\gamma$ controls the level of noise provided into the gradients in the outer or meta loop during the learning phase. No co-learner component is used if the value of $\gamma$ is 0. Different values for $\gamma$ were used to study the noise intensity. The used values were 0.2, 0.4, 0.6, 0.8, and 1.0. The performance of the effect of $\gamma$ is shown in Fig.~\ref{gamma_perf}.

The performance analysis in Fig.~\ref{gamma_perf} reveals that the co-learner's effectiveness highly depends on the intensity of noise introduced via the variable $\gamma$. When $\gamma$ is set to 0.2, the model achieves the highest accuracy and F1 scores in 5-way 1-shot and 5-way 5-shot settings, indicating optimal gradient regularization that enhances generalization without overloading the learning process. However, performance declines significantly as $\gamma$ increases beyond this value. The reduction in performance is caused by high label noise, which confuses the model and leads to incorrect learning. Excessive noise also disrupts the learning process, slowing convergence or causing divergence. Both factors show the importance of managing noise levels for stable training~\citep{karimi2020deep, jim1994effects, ding2022impact}.

\subsection{Results on Cross Validation}
\label{fsl_cross_validation}
To evaluate the generalization capability and robustness of CCoMAML, experiments were conducted using two disjoint splits of the muzzle dataset. In the first setting, UNE muzzle data is used for training, and the UNL muzzle data for testing. To further assess cross-domain generalization, the splits were reversed in the second setting (training on UNL muzzle and testing on UNE muzzle). This approach ensures that training and testing classes remain disjoint, allowing for a fair evaluation of the CCoMAML's ability to adapt to previously unseen data distributions. 
This setup also reflects real-world scenarios involving dynamically changing herd sizes. When new cattle are added to the herd, the size increases; when some are removed, the size decreases. The experiment, therefore, evaluates the proposed model’s performance under varying herd sizes.

\begin{figure}[ht]
\centering
\includegraphics[width=0.98\textwidth]{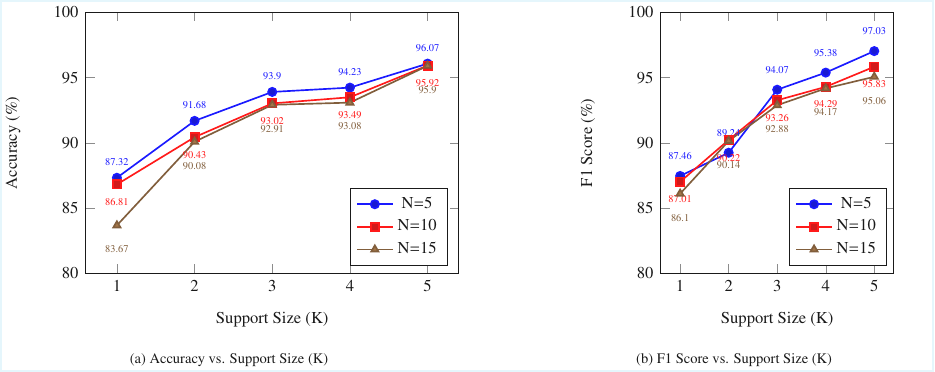}
\caption{N-way K-shot performance of CCoMAML using UNE muzzle as training and UNL muzzle as testing dataset.}
\label{fsl_une}
\end{figure}

\begin{figure}[ht]
\centering
\includegraphics[width=0.98\textwidth]{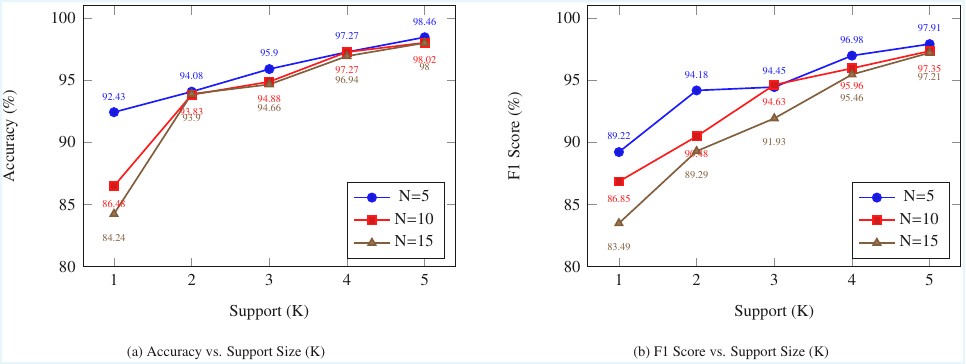}
\caption{Performance of CCoMAML on N-way K-shot classification using UNL muzzle as training data and UNE muzzle as testing data.}
\label{fsl_unl}
\end{figure}

The evaluation follows an N-way K-shot classification framework, a standard practice in few-shot learning. The number of classes \(N\) is set to 5, 10, and 15, while the number of samples per class \(K\) varies as 5, 4, 3, 2, and 1. We selected \(K\) up to 5 because the minimum number of images per cattle in the UNE dataset is 6. In the case of the UNL dataset, 12 classes contain fewer than 6 images per cattle, making them unsuitable for the 5-shot setting. Therefore, only the remaining 256 classes were used for evaluation. The results for both dataset splits are presented in Fig.~\ref{fsl_une} and Fig.~\ref{fsl_unl}.

The experimental results in Figures ~\ref{fsl_unl} and ~\ref{fsl_une} demonstrate strong cross-dataset generalization in few-shot learning (FSL) for cattle identification using muzzle images from the UNL and UNE datasets. Performance consistently improves as the number of shots (K) increases, indicating that the model benefits from more training examples per class. The best performance is achieved at 5-way 5-shot (N=5, K=5) with 98.46\% accuracy and 97.91\% F1-score, highlighting the model’s effectiveness under well-supported conditions.

However, accuracy and F1-score decline as the number of classes (N) increases and the number of shots (K) decreases. This trend reflects the increased difficulty of discriminating among more classes with limited examples, consistent with established theoretical findings~\citep{luo2023closer, cao2019theoretical}. The lowest performance, observed at 15-way 1-shot (N=15, K=1) with 84.24\% accuracy, underscores the challenge of low-shot learning.

Overall, the consistent performance trends and strong results across dataset swaps confirm the model’s robustness and its ability to generalize across different but related datasets in a cross-validation setting.

\section{Evaluating DL Models with Few Training Samples}
\label{DLvsFSL}
\begin{figure}[ht]
\centering
\includegraphics[width=0.98\textwidth]{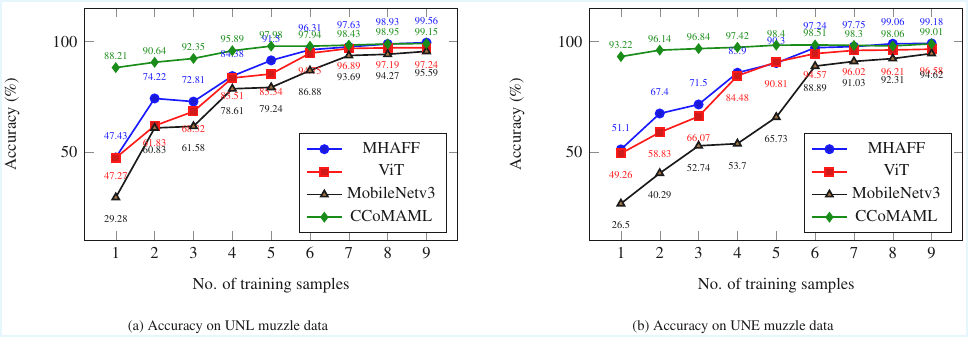}
   \caption{Comparison of accuracy vs. number of training images for UNL and UNE muzzle data across different DL models and CCoMAML}
    \label{DLvsFSLcompare}
\end{figure}

The objective of this section is to evaluate and compare the performance of DL models with the CCoMAML few-shot learning approach under conditions of limited training data (i.e., fewer than ten muzzle images per class). This enables a fair and consistent comparison with CCoMAML, which is specifically designed for low-data scenarios. The three DL models—MHAFF, ViT, and MobileNetV3 are selected based on their superior performance across both the UNL and UNE datasets~\citep{dulal2025mhaff}. 
From the UNE and UNL muzzle datasets, only those classes containing at least 10 images were selected. Exactly 10 images per class were uniformly sampled. As a result, the UNE muzzle dataset contains 65 classes, and the UNL muzzle dataset contains 227 classes. The models are then trained using $K$-shot learning, where $K \in \{1, 2, \ldots, 9\}$. Classification accuracy is used to assess model performance, as shown in Fig.~\ref{DLvsFSLcompare}. The results for CCoMAML are based on 5-way $K \in \{1, 2, \ldots, 9\}$ shots. 

Figure~\ref{DLvsFSLcompare} provides an extended comparative analysis of model performance across a broader range of training sample sizes ($K = 1$ to $9$) for both the UNL and UNE muzzle datasets. All models exhibit increasing classification accuracy as the number of training samples per class grows, reflecting improved generalization with more data.

Among the deep learning baselines, MHAFF consistently outperforms ViT and MobileNet-v3 across nearly all support sizes on both datasets, reaffirming its robustness in limited-data scenarios. MobileNet-v3 remains the weakest performer, particularly under extremely low-shot conditions (e.g., 29.28\% on UNL and 26.5\% on UNE at $K = 1$), indicating its limited capacity to generalize from sparse samples. By contrast, MHAFF achieves substantial gains at higher $K$ values, reaching 99.56\% accuracy on UNL and 99.18\% on UNE at $K = 9$.

The use of CCoMAML further highlights the effectiveness of few-shot meta-learning. Notably, it maintains high accuracy even with minimal training data, achieving 88.21\% and 93.22\% accuracy at $K = 1$ on UNL and UNE, respectively. The performance advantage is especially pronounced in low-data regimes, where standard deep learning models show steep drops in accuracy, while CCoMAML remains stable.

Across both datasets, when the number of training samples reaches 7 or more, all selected deep learning models consistently achieve over 90\% accuracy. In such cases, transfer learning using ImageNet-pretrained weights also yields satisfactory performance. However, the key distinction emerges in extremely low-data settings—for instance, with only 3 training samples—where all transfer learning-based models fall below 75\% accuracy, whereas CCoMAML exceeds 90\%. When limited to just a single training sample, all three baseline deep learning models perform poorly, while CCoMAML continues to deliver robust performance. These finding shows the potential of few-shot learning approaches such as CCoMAML in data-constrained environments. Its superior ability to adapt quickly with minimal supervision makes it well-suited for practical applications like cattle identification, where collecting large annotated datasets may be infeasible.


\section{Conclusion}
\label{conclusion}
This study introduced CCoMAML, a novel meta-learning framework for individual cattle identification using muzzle images, specifically designed to perform well even with limited data and dynamically changing herd sizes. Unlike conventional methods, CCoMAML eliminates the need for retraining from scratch when new cattle are added, making it highly practical for real-world applications.

A key improvement of this work is the CNN-based co-learner, which enhances model generalization by modifying gradient updates through learnable noise injection. This mechanism enables more robust learning and better adaptation to new tasks with minimal data.

A comprehensive evaluation against 14 state-of-the-art few-shot learning (FSL) methods was conducted, making this the first study to systematically benchmark FSL techniques for cattle identification based on muzzle patterns. The proposed model achieved a top accuracy of 98.46\% and F1-score of 97.91\% in a 5-way 5-shot setting, significantly outperforming existing methods.

In addition, this research rigorously explored the minimum number of images required for effective identification using various N-way K-shot settings, demonstrating that performance improves with more shots per class but declines with an increase in the number of classes. These findings highlight the scalability and robustness of CCoMAML under diverse and realistic conditions.

In summary, CCoMAML presents a powerful, adaptive, and data-efficient solution for cattle identification, paving the way for more intelligent and scalable livestock management systems.

\section*{Acknowledgment}
This project was supported by funding from Food Agility CRC Ltd, funded under the Commonwealth Government CRC Program. The CRC Program supports industry-led collaborations between industry, researchers, and the community. We also thank Dr Shawn McGrath, Mr. Jonathan Medway, and Prof. Dave Swain for their assistance with the project.
\bibliographystyle{elsarticle-harv}
\bibliography{reference}

\end{document}